\documentclass[12pt]{article}

\usepackage{newtxtext,newtxmath}
\usepackage{graphicx}
\usepackage[letterpaper,margin=1in]{geometry}

\renewenvironment{abstract}
{\quotation}
{\endquotation}

\date{}

\makeatletter
\renewcommand{\fnum@figure}{\textbf{Figure \thefigure}}
\renewcommand{\fnum@table}{\textbf{Table \thetable}}
\makeatother

\usepackage{scicite}

\usepackage{url}
\usepackage{array}      
\usepackage{tabularx}  
\usepackage{booktabs}
\usepackage{diagbox}
\usepackage{graphicx}
\usepackage{lineno,hyperref}
\usepackage{subfig}
\usepackage{subfloat}
\usepackage{setspace}

\newtheorem{theorem}{Theorem}{}

\makeatletter
\renewcommand{\fnum@figure}{\textbf{Fig. \thefigure}}
\renewcommand{\fnum@table}{\textbf{Table \thetable}}
\makeatother
{}
\graphicspath{{figuresSM/}}

\newcounter{moviecounter}

\def\scititle{
	Self-Organizing Aerial Swarm Robotics for Resilient Load Transportation :
	A Table-Mechanics-Inspired Approach
}
\title{\bfseries \boldmath \scititle}

\author{
	Quan Quan$^{1\ast\dagger}$,
	Jiwen Xu$^{2,1\dagger}$,
	Runxiao Liu$^{1\dagger}$,
	Yi Ding$^{1}$,
	Jiaxing Che$^{1}$,
	Kai-Yuan Cai$^{1}$\and
	\small$^{1}$School of Automation Science and Electrical Engineering, Beihang University, Beijing 100191, China.\and
	\small$^{2}$Department of Mechanical and Automation Engineering, The Chinese University of Hong Kong,\and
	\small Hong Kong SAR 999077, China.\and
	\small$^\ast$Corresponding author. Email: qq$\_$buaa@buaa.edu.cn\and
	\small$^\dagger$These authors contributed equally to this work.
}

\begin{document} 
	
	\maketitle
	
	\begin{abstract} \bfseries \boldmath
		In comparison with existing approaches, which struggle with scalability, communication dependency, and robustness against dynamic failures, cooperative aerial transportation via robot swarms holds transformative potential for logistics and disaster response. Here, we present a physics-inspired cooperative transportation approach for flying robot swarms that imitates the dissipative mechanics of table-leg load distribution. By developing a decentralized dissipative force model, our approach enables autonomous formation stabilization and adaptive load allocation without the requirement of explicit communication. Based on local neighbor robots and the suspended payload, each robot dynamically adjusts its position. This is similar to energy-dissipating table leg reactions. The stability of the resultant control system is rigorously proved. Simulations demonstrate that the tracking errors of the proposed approach are 20${\%}$, 68$\mathbf{\%}$, 55.5$\mathbf{\%}$, and 21.9$\mathbf{\%}$ of existing approaches under the cases of capability variation, cable uncertainty, limited vision, and payload variation, respectively. In real-world experiments with six flying robots, the cooperative aerial transportation system achieved a 94$\mathbf{\%}$ success rate under single-robot failure, disconnection events, 25$\mathbf{\%}$ payload variation, and 40$\mathbf{\%}$ cable length uncertainty, demonstrating strong robustness under outdoor winds up to Beaufort scale 4. Overall, this physics-inspired approach bridges swarm intelligence and mechanical stability principles, offering a scalable framework for heterogeneous aerial systems to collectively handle complex transportation tasks in communication-constrained environments.
	\end{abstract}
	
	\section*{Introduction}
	
	Load transport is a fundamental component of modern society. With the advent of diverse aerial vehicles \cite{vasarhelyi2018optimized, mcguire2019minimal, o2022neural, zhou2022swarm, ren2025safety, kaya2025aerial, farivarnejad2022multirobot}, aerial transportation has emerged as a versatile solution for the transportation of cargo, transcending the constraints of geographical limitations and overcoming the limitations of ground and water transportation. However, in contrast to ground robots, which benefit from the capacity to transport heavy payloads over extended distances, aerial robots are constrained by thrust limitations. This has imposed strong limitations on their potential applications \cite{oishi2022cooperative}. Given the considerable expense and technical difficulties imposed on the development of large-scale flying robots, the use of multiple regular-sized flying robots is being considered as a potential solution for efficient transportation \cite{loianno2017cooperative, jackson2020scalable}. Several approaches to the cooperative transportation of flying robots are currently being explored \cite{rao2023path}. The rigid attachment has been a commonly used solution for addressing bar-shaped loads with two flying robots \cite{chen2021cooperative, wu2023cooperative,oishi2021autonomous,mu2019universal}. The flying robots themselves can adjust the flight attitude of the load, but this introduces additional interference, which jeopardizes the stability of the robots. In parallel, soft attachment with cable addresses the load transportation, and the solution is insensitive to the flying attitude of robots. Furthermore, the low cost and convenient deployment of this soft attachment solution have led to increasing interest in working on cooperative transportation with flying robots \cite{sreenath2013trajectory, zhang2023formation, li2023rotortm, wang2024robust}. However, cooperative transportation continues to face significant difficulties. Unexpected communication interference poses a major obstacle to achieving precise formation control, while complex weather conditions add further uncertainty when operating in dynamic environments. Furthermore, mid-flight failures may occur, potentially jeopardizing flight safety \cite{yu2018distributed}. These practical scenarios and potential issues drive a pressing need to develop a robust and generalized transportation framework to address these difficulties and ensure the safety of cooperative transport operations.
	
	\paragraph*{Why Cooperative Transportation of Cable Suspended Payloads with Aerial Robotics}
	
	Intuitively, there are two primary approaches to increasing the payload capacity of transportation robots. The first is to design a larger aerial robot capable of carrying heavier loads, while the second involves coordinating multiple smaller aerial robots to work collaboratively for transport \cite{liu2024coordinated,tagliabue2019robust}. Although system complexity rises with the number of robots, the latter approach offers several significant advantages over the former and exhibits the following key features.
	
	(i) \textbf{Scalable}. The payload capacity can be readily scaled by increasing the number of aerial robots. This cooperative strategy is commonly observed in nature, where individuals work together to transport food or nesting materials in a flexible manner, free from the constraints of a fixed number, formation, or group composition \cite{feinerman2018physics}. Individuals can join or leave a transport mission without requiring the payload to be reallocated, significantly enhancing transportation efficiency. Moreover, this approach greatly reduces costs associated with design, testing, production, and maintenance\cite{qian2022robust}.
	Additionally, the design process for flying robots with varying sizes and payload capacities can be streamlined, eliminating the need for repeated design iterations for different transportation tasks. Compared to a single large robot carrying heavy objects, a group of lightweight, small robots working together to transport heavy loads is more consistent with natural phenomena and evolutionary trends.
	
	(ii) \textbf{Robust}. In a cooperative transportation mechanism, attaching multiple components to the payload enhances both the vehicle's stability and its control \cite{wang2024robust}. The deployment of multiple flying robots inherently introduces redundancy, significantly improving the transportation process's overall safety and reliability. In the event of multiple robot failures, it is crucial that the remaining robots adapt to the sudden changes and continue to perform their mission. Compared to a single-robot scenario, a cooperative strategy enables a more robust and resilient transport process \cite{liang2019novel,tagliabue2019robust}, and the over-actuated nature of the cooperative transportation system allows multiple robots to stably control the suspended load in terms of both position and attitude, ensuring enhanced stability and safety throughout the operation.
	
	(iii) \textbf{Flexible}. When transporting loads outdoors, complex environments such as narrow terrains often require deformable formations to ensure safe passage. A cooperative transportation mechanism is well-suited to meet such demands due to the flexibility in both the number of robots and their formation. Additionally, cooperative strategies enable the implementation of various flying modes to accommodate different tasks. In scenarios where the load is deformable or has an irregular shape, a cooperative transportation mechanism can adjust the suspending forces dynamically, ensuring a stable attitude for the suspended load. Different transportation structures are displayed in Table~\ref{tbl:training1}.	
	\begin{table}[!htb]
		\centering
		\caption{\textbf{Comparison of different transportation structures.} Rigid attachment v.s. soft attachment}
		\label{tbl:training1}
		\renewcommand\arraystretch{1.2}
		\normalsize
		\begin{tabularx}{\linewidth}{>{\centering\arraybackslash}X>{\centering\arraybackslash}X>{\centering\arraybackslash}X}
			\toprule[1pt]
			\textbf{Type} & \textbf{Rigid} & \textbf{Soft} \\
			\midrule
			Connector & Gripper, Manipulator, Magnets & Cable \\
			Advantages & Grasp ability, avoiding undesired oscillations, more actuated degrees of freedom & Low cost, lightweight, safety, fast deployment \\
			Drawbacks & Difficult to design, more energy-consuming for active grippers, additional inertia for payload & Inability to exert negative (pushing) forces, fewer actuated degrees of freedom, undesired oscillations \\
			\bottomrule[1pt]
		\end{tabularx}
	\end{table}
	
	\paragraph*{Why No explicit communication}
	A common prerequisite for collaborative robot transportation is a dependable, low-latency communication channel. This is not an issue for ground robots, as they are in close proximity and have the capacity to transport ``heavy'' high-performance communication equipment, which can enhance communication bandwidth, mitigate communication latency, and reduce failures \cite{huzaefa2021force}.
	However, due to the limited payload capacity of aerial robots, they face challenges in establishing communication environments comparable to those of ground robots. As a result, performing trajectory planning independently on the onboard computing device of each aerial robot presents a viable solution\cite{wang2018cooperative,geng2020cooperative,liu2021analysis}. The cooperative transport strategy, which does not rely on explicit communication, is usually based on the leader-follower cooperative paradigm \cite{tagliabue2017collaborative,wang2016force,gassner2017dynamic}.  Each aerial robot then follows its own planned trajectory. However, this approach places high demands on the homogeneity of the flying robots, as both the load distribution and trajectory planning primarily follow the principle of equal distribution. This is not easy to achieve in flying robots with different capabilities and limitations, particularly when the leader fails. A fixed formation of robots is also a viable solution for cooperative transportation, where the problem of cooperative transport is transformed into a formation control problem \cite{abeshtan2024robust,shirani2019cooperative,hegde2022multi,meissen2017passivity}. In this framework, the flying robots only need to maintain an appropriate distance from their neighbors to sustain the formation. Nevertheless, extending this method to an arbitrary number of flying robots is not straightforward, and the aforementioned approaches are not easily applicable to aerial robots with varying capabilities and limitations. The pre-designed formations of general solutions impede the flexibility and scalability of aerial robots, rendering them susceptible to failure during the procedure. Admittance control offers an alternative solution for cooperative transportation without the need for explicit communication  \cite{augugliaro2013admittance,romano2022cooperative}. By building upon the leader-follower framework, the follower robot estimates the external forces acting on it using only its model and the onboard inertial measurement unit (IMU). Based on this estimation, the robot then adjusts its dynamics to follow a reference trajectory, which is dynamically modified by the admittance controller according to both the estimated external force and the desired trajectory. This method is particularly applicable to a swarm of follower robots with varying capabilities and limitations, as well as in scenarios involving human-robot cooperation \cite{yu2021adaptive}. The flexibility of this approach makes it suitable for a wide range of applications where robots need to respond to external forces while maintaining their collective mission objectives. However, it is important to note that admittance control requires a leader to guide the robot swarm, and it is inapplicable in situations where the cable is slack or relaxed. In such cases, the system's stability and coordination could be compromised, limiting its effectiveness. Performance analysis of different methods is demonstrated in Table~\ref{tbl:training2}. 

	\begin{table}[!htb]
		\normalsize
		\centering
		\caption{\textbf{Performance analysis of different methods.} ``+'' represents strength/requirement level.}
		\label{tbl:training2}
		\renewcommand\arraystretch{1.2}
		\resizebox{\linewidth}{!}{
			\begin{tabular}{ccccc}
				\toprule[1pt]
				\diagbox{Method}{Performance} & Scalability & Robustness & Communication Requirement & Fault Tolerance\\
				\midrule
				Formation-based\cite{meissen2017passivity} & + + & + & + & + \\
				Payload Leader\cite{ping2023collaborative} & + & + & + & +\\
				Admittance Control\cite{romano2022cooperative} & + + & + & + & + +\\
				Dissipative Control & + + + & + + + & + & + + + \\
				\bottomrule[1pt]
			\end{tabular}
		}
	\end{table}

	Overall, in complex environments, maintaining explicit communication between the flying robots is not always feasible, leading to non-trivial instabilities and coordination issues\cite{javaid2023communication}. As the number of flying robots increases, the challenge of efficiently allocating and coordinating flying robots becomes a significant hurdle during the cooperative transportation process.  Additionally, halfway failures, where individual robots may experience malfunctions or unexpected behavior, further complicate the cooperative process. These factors—load allocation, communication limitations, halfway failures, and environmental complexities—are crucial considerations in the design of a scalable, robust, and communication-free transport framework for a swarm of flying robots. Addressing these challenges is key to ensuring reliable and efficient multi-robot cooperation in real-world applications.
	
\begin{figure}[!htbp]
		\centering
		\includegraphics[width=0.9\linewidth]{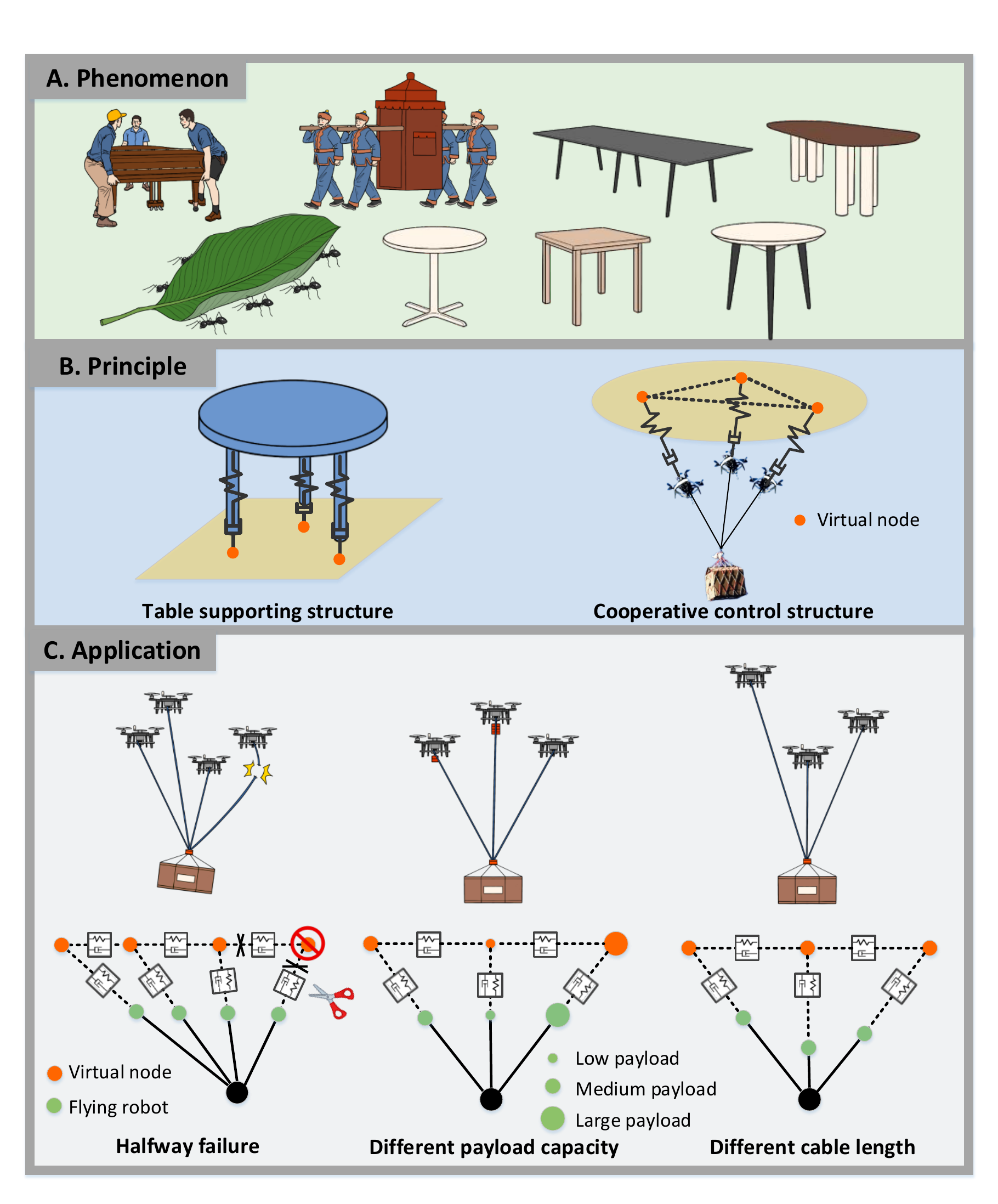}\\
		\caption{\textbf{Phenomenon, principle, and application of cooperative transportation.} (\textbf{A}) Cooperative transport framework inspired by social and natural transport phenomena, and adaptive load distribution based on table structure. (\textbf{B}) Table supporting structure and a demonstration of three flying robots carrying a suspended load based on the design of virtual nodes. (\textbf{C}) Different cases of the dissipative cooperative transportation system. Green dots represent the flying robots, while orange dots denote their corresponding virtual nodes. The dotted lines illustrate the dissipation forces.}
		\label{fig:introduction}
	\end{figure}

	\paragraph*{What can be learned from tables} To this end, we put forth a dissipative system as a flexible yet robust transport paradigm. The ``table leg support problem'', as proposed by Euler in his 1774 dissertation at the St. Petersburg Academy of Sciences, aims to determine the four reaction forces of the legs of a rectangular table when a weight is placed on its surface \cite{euler1774pressione}. This problem is viewed as a redundant control problem, as a table supported by four legs has only three degrees of freedom—pitch, roll, and height—when considered as a static system. With four or more supporting table legs, the ``smart'' table automatically determines four or more reaction forces based on the load weight. Such cooperation structures are frequently observed in nature in Fig.~\ref{fig:introduction}A, and in all possible cases, the one that occurs is the one that uses the least energy. Euler proposed treating the supporting surface as a soft ground and introduced the concept of elastic deformation so that the solution for rigid support could be derived as a limiting case of the soft support scenario \cite{mu2019universal}. As illustrated in Fig.~\ref{fig:introduction}B, table legs constitute a spring-damper structure. The more evenly the legs are distributed around the table's center of gravity, the more stable the table becomes. Even in the event of a failure of one or more legs, this balanced configuration helps maintain stability and prevents the table from tipping over. Based on this fundamental understanding of the dissipative structure of tables, a dissipative transportation paradigm is developed that allows the flying robots to be self-organized for resilient load transportation.
	
	In this paper, the cooperative transportation is constructed as a dissipative system with flying robots, virtual nodes, and a suspended load as in Fig.~\ref{fig:introduction}B. Dissipative force is generated among virtual nodes and between virtual nodes and flying robots, based on local information from their neighbors. 
	As illustrated in Fig.~\ref{fig:introduction}C, we demonstrate a two-dimensional dissipative transport system. This framework enables adaptive load distribution and remains robust to both robot failure and dynamic reconfiguration, such as the entry or exit of individual flying robots during transport. In the case of halfway failure, the formation autonomously reconfigures through local dissipative interactions among the remaining three flying robots, restoring equilibrium without external intervention. When a new robot joins during transport, the newly involved robot is connected to the payload and its neighbors, enabling the system to stabilize into a new equilibrium state. For heterogeneous payload distributions, the stability of the suspended load can be regulated by tuning the coefficient of virtual ``spring'', and the dissipative interaction mechanism enhances system robustness by tolerating uncertainties in cable length.
	
	Building on the above demonstration, this table-mechanics-inspired system flexibly adapts to variations in robot team size and maintains performance in the event of mid-flight failure.  To the best of our knowledge, this is a collaborative transport framework that simultaneously addresses cable length uncertainty, communication discontinuity, potential robot failure, and supports heterogeneous robots in {\it{the first time}}, which are key challenges for real-world deployment.
	
	\section*{Results}
	\begin{figure}[!htbp]
		\centering
		\includegraphics[width=0.9\linewidth]{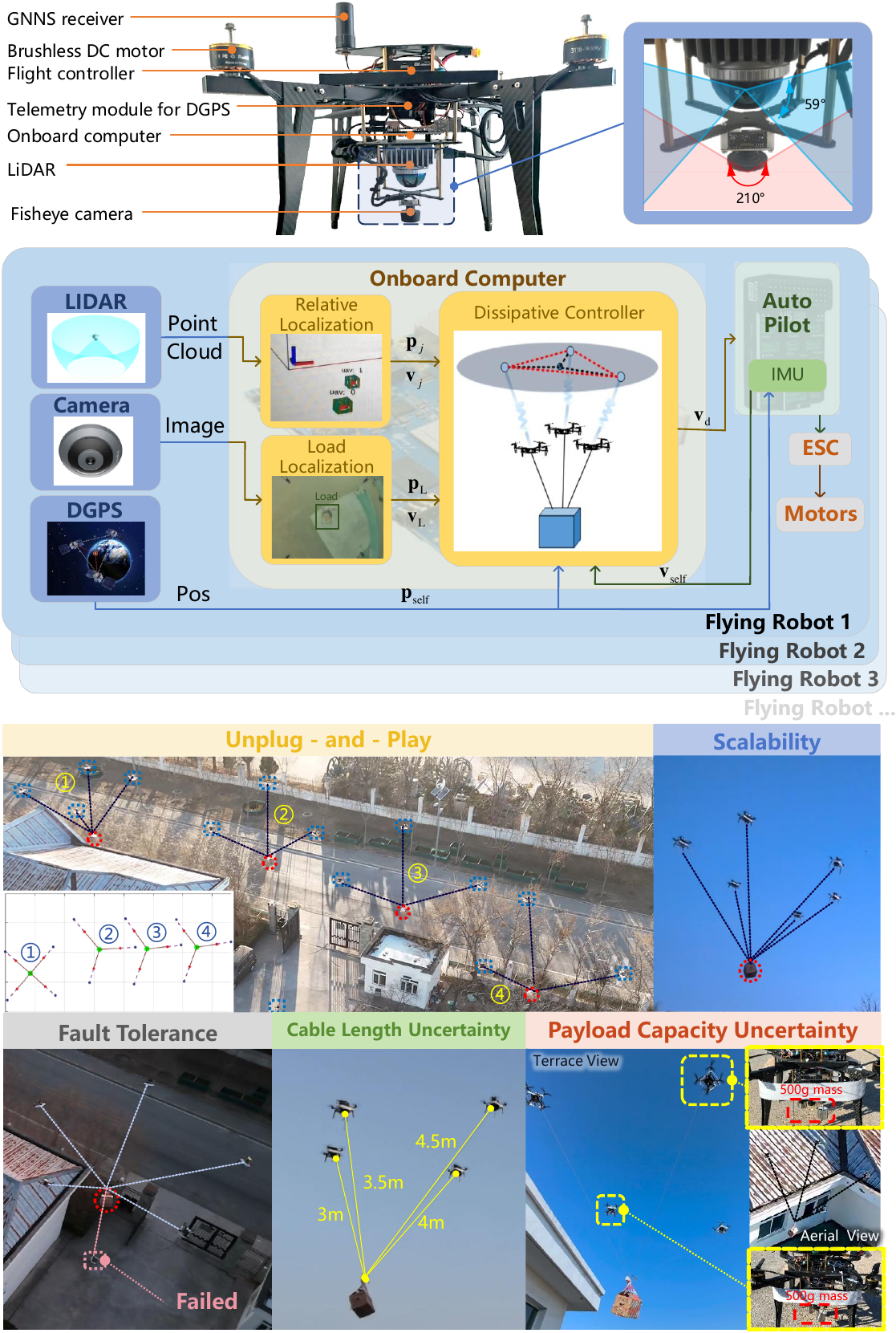}\\
		\caption{\textbf{Configuration and capabilities of flying robots.} The hardware configuration, multi-robot framework, and transportation challenges are demonstrated.}
		\label{fig:UAV1}
	\end{figure}
	The proposed cooperative transportation system is validated through a combination of simulations and real-world flight experiments. In physical tests, we implement a cooperative aerial transport system composed of flying robots, suspension cables, hooks, and payloads. 
	Extensive outdoor experiments are conducted to evaluate the proposed transportation system under diverse scenarios, including robustness to cable length uncertainties, resilience to heterogeneous payload capacities, and the ability to dynamically remove flying robots during transport.
	
	
	\paragraph*{Five Flying Robots Cooperative Transport}
	
	\begin{figure}[!htb]
		\centering
		\includegraphics[width=\linewidth]{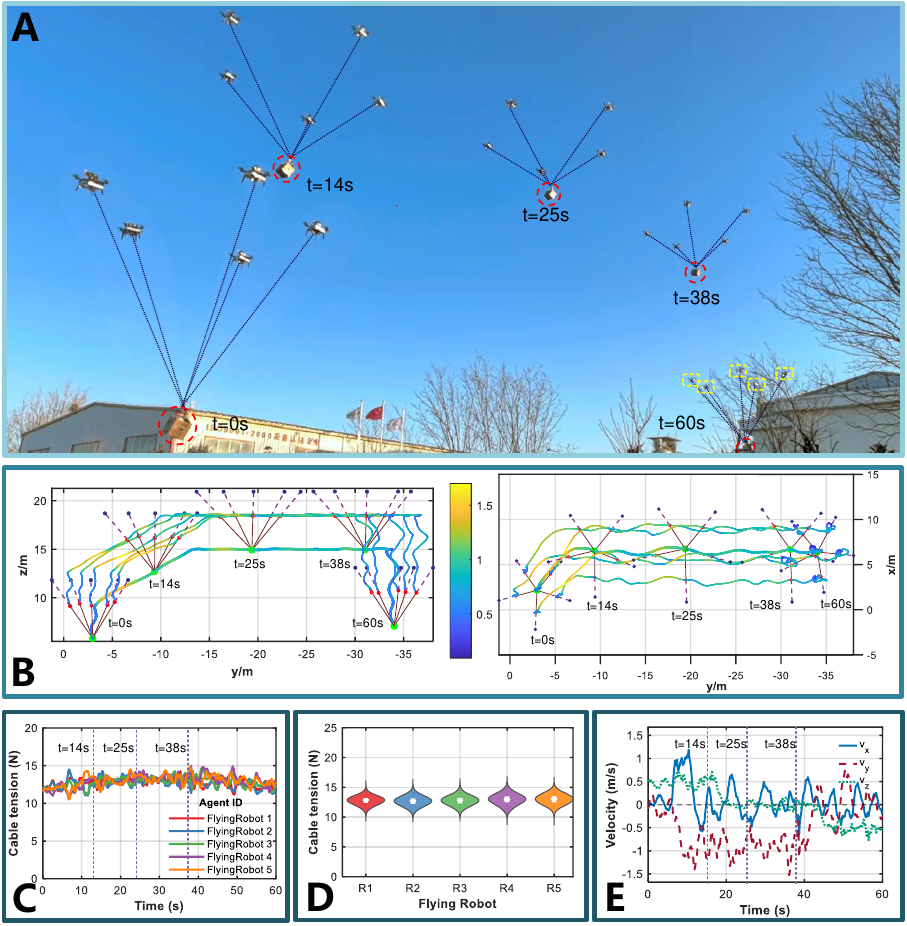}\\
		\caption{\textbf{Five flying robots cooperative transport.} (\textbf{A}) Positions and states of the flying robots, cables, and load at representative moments during  the transportation. (\textbf{B}) Positions of virtual nodes, flying robots and the load at representative moments during the transportation. Trajectories of the flying robots and the load from different views. (\textbf{C}) The change of cable tension of each flying robot over time. (\textbf{D}) The distribution of the cable tensions during the whole transportation task. (\textbf{E}) The change of the velocity of one flying robot over time. (See Movie S1 in Supplementary Materials)}
		\label{fig:Five_UAV}
	\end{figure}
	In real-world flight experiments, we evaluate the transport system with teams of four, five, and six flying robots tasked with delivering a payload to a target location. A detailed analysis of the experiment with five flying robots is provided.  Here, we focus on a representative scenario involving five aerial robots collaboratively transporting a 5 kg payload. After taking off, the load is taken to a height of 15 m above the ground and is precisely delivered to a second-floor terrace located approximately 34 m away and 5 m above the initial takeoff point.
	The experimental results are shown in Fig.~\ref{fig:Five_UAV}. Fig.~\ref{fig:Five_UAV}A illustrates the positions and states of the flying robots, cables, and load at representative moments throughout the transport sequence. Cables are specifically marked with black dashed lines. Fig.~\ref{fig:Five_UAV}B presents the state of the system and load trajectories at five representative moments. Large green dots indicate the payload, red dots represent the flying robots, and dark blue dots denote the virtual nodes associated with each robot. Trajectory colors encode velocity, with color gradients mapped to the speed of each object. Under the proposed method, the spatial distribution of flying robots and their corresponding virtual nodes remains well-organized, exhibiting near symmetry throughout the transport task. 
	The speed of the load and the flying robots remains close to 1 m/s, with a maximum speed reaching 1.6 m/s.
	Fig.~\ref{fig:Five_UAV}C and Fig.~\ref{fig:Five_UAV}D illustrate the temporal variation in cable tension for each flying robot, along with a violin plot showing the overall tension distribution. The results indicate that the load is supported with a balanced and well-distributed allocation of tension among the robots. Fig.~\ref{fig:Five_UAV}E shows the velocity profile of a representative flying robot during transport, demonstrating relatively stable speed throughout the task. The black dashed lines in Fig.~\ref{fig:Five_UAV}C and Fig.~\ref{fig:Five_UAV}D mark the five key time points referenced earlier. Despite noticeable external wind disturbances (approximately Beaufort scale 4), as evidenced by the flag motion in Fig.~\ref{fig:Five_UAV}A, which introduced fluctuations in both robot trajectories and cable tensions, the system maintained stability and successfully completed the cooperative transport task with high precision.
	
	\paragraph*{Flying Robots with Different/Uncertain Cable Lengths}
	
	\begin{figure}[!htb]
		\centering
		\includegraphics[width=\linewidth]{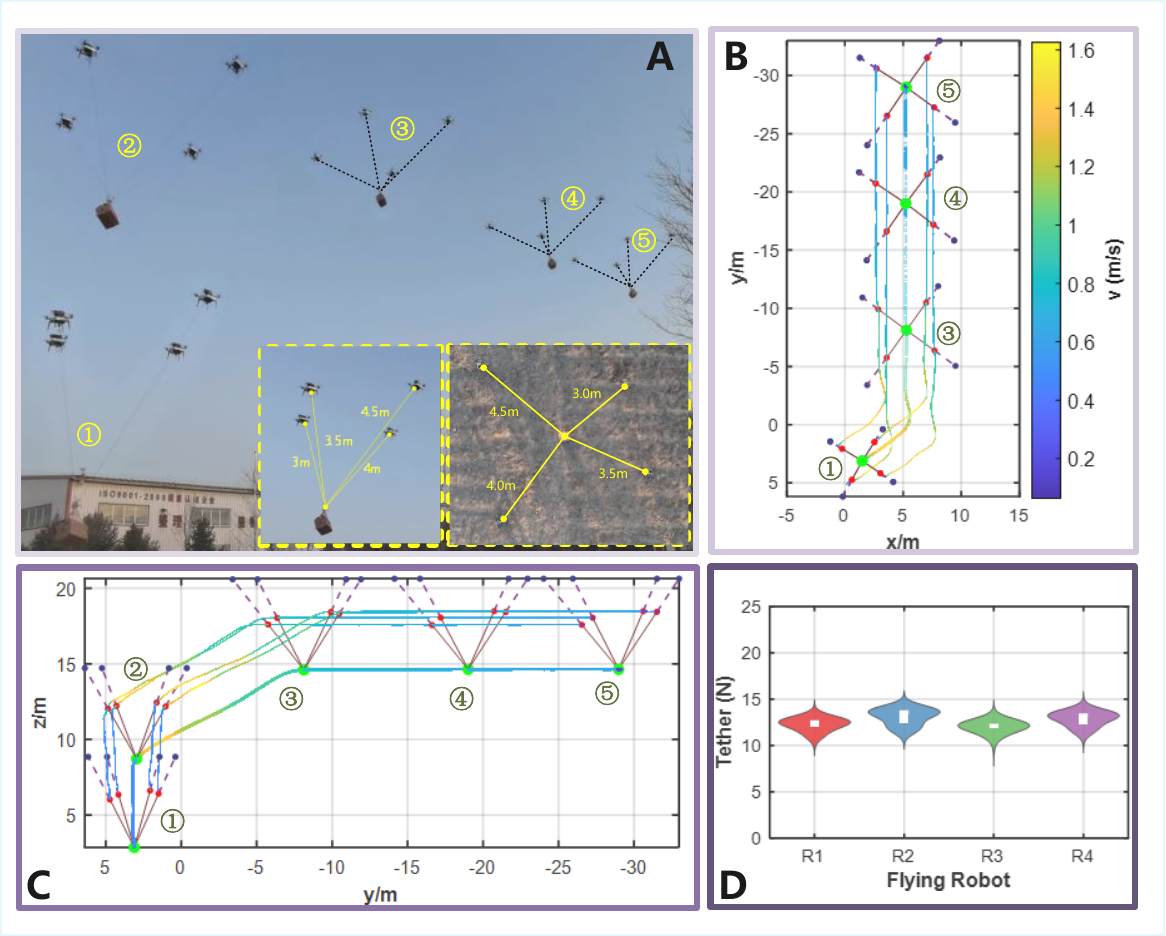}\\
		\caption{\textbf{Real-world flight Experiment with different cable lengths.} (\textbf{A})  Positions and states of the flying robots, cables, and load at representative moments during the transportation. The lengths of the cables varies from 3.0 m  to 4.5m. (\textbf{B}) The top view of positions and trajectories of the virtual nodes, flying robots and the load. (\textbf{C}) The side view of positions and trajectories of the virtual nodes, flying robots and the load.  (\textbf{D}) The distribution of the cable tensions during the whole transportation task.
		(See Movie S2 in Supplementary Materials)}
		\label{fig:Cable_length}
	\end{figure}

	This real-world flight experiment demonstrates the robustness of the proposed system in the presence of varying or uncertain cable lengths. In practical transport scenarios, ensuring uniform or precisely measured cable lengths is often infeasible. We introduced a flying robot with an uncertain cable length into the cooperative transportation task during the experiment. In this experiment, four flying robots transport a 4 kg payload. As shown in Fig.~\ref{fig:Cable_length}A, the lengths of the cables are 3 m, 3.5 m, 4 m, and 4.5 m, respectively. Referring to Fig.~\ref{fig:Cable_length}B and Fig.~\ref{fig:Cable_length}C, the state of the transport system is visualized throughout the transport process. The large green dot represents the payload, the red dots indicate the flying robots, and the dark blue dots denote the corresponding virtual nodes. Despite the substantial differences in cable lengths, the proposed method enables the flying robots to autonomously adjust their altitude and orientation, maintaining cable tautness and achieving a balanced distribution of tension. As shown in the violin plot in Fig.~\ref{fig:Cable_length}D, although external wind disturbances introduce fluctuations and slight differences in the mean tension among the four flying robots, the overall distribution remains relatively uniform. This result validates the effectiveness of the proposed method in handling cable length variability. Despite the differences in cable lengths, the flying robots adaptively adjust their positions relative to one another and to the payload, ensuring appropriate load allocation throughout the transport process. 
	due to the field-of-view limitations of the LiDAR sensor (-7° to 53°), 
	large discrepancies in cable lengths can lead to significant altitude differences among the flying robots, potentially disrupting relative localization. In such cases, incorporating a communication-based method to obtain relative position information may be necessary to ensure stable system operation.

\begin{figure}[h!tb]
		\hspace{-3mm}
		\includegraphics[width=1\linewidth]{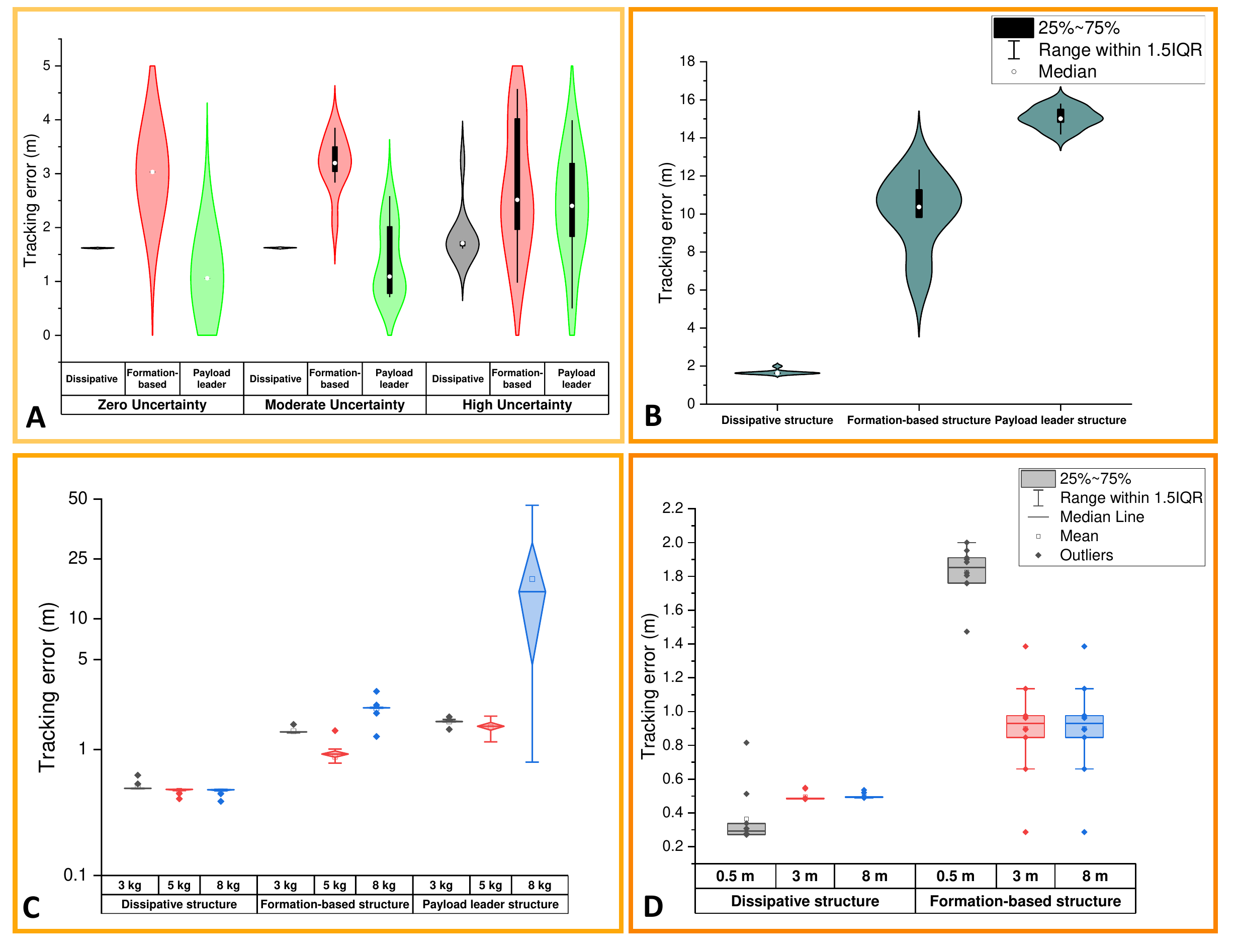}\\
		\caption{\textbf{Simulation benchmark of dissipative structure, payload leader structure, and formation-based structure.} (\textbf{A}) Tracking error distributions of the proposed dissipative structure, formation-based structure, and payload leader structure across different cable length uncertainties. The average of cable length is 2.02 m, and uncertainty is modeled as 0, a variable in a uniform distribution from -0.5 m to 0.5 m, and from -1.01 m to 1.01 m,  respectively. (\textbf{B}) Tracking error distributions of the proposed dissipative structure, formation-based structure, and payload leader structure across different capabilities. The payload of flying robots is selected from a uniform distribution from 11 N to 19 N. (\textbf{C}) Tracking error distributions of the proposed dissipative, formation-based, and payload leader structures across different loads. The load is selected as 3 kg, 5 kg, and 8 kg. (\textbf{D}) Tracking error distributions under different perception ranges for dissipative, formation-based, and payload leader structures. Each experiment was repeated 10 times to extract statistical properties. A movie of comprehensive simulation is presented in Movie S6 in Supplementary Materials.}
		\label{fig5}
	\end{figure}

	The violin plot in Fig.~\ref{fig5}A shows the distribution of load tracking errors for 10 flying robots operating under varying cable length uncertainties. For the proposed method (grey), the median tracking error remains consistently around 1.6 m, with the most compact distribution among the three methods. In contrast, the formation-based (red) and payload leader (green) methods exhibit significant increases in tracking error as uncertainty grows. Specifically, under high uncertainty, the median tracking errors for both the formation-based and payload leader methods are approximately double compared to the dissipative method, with wider distributions and greater variability. The dissipative structure maintains stability by anchoring virtual nodes at fixed altitudes, thereby offering greater tolerance to cable length fluctuations. A detailed analysis of the performance under different uncertainty levels is provided in Section S6 of Supplementary Materials.
	\paragraph*{Flying Robots with Different/Uncertain Payload Capacities}
	
	\begin{figure}[htbp]
		\centering
		\includegraphics[width=0.8\linewidth]{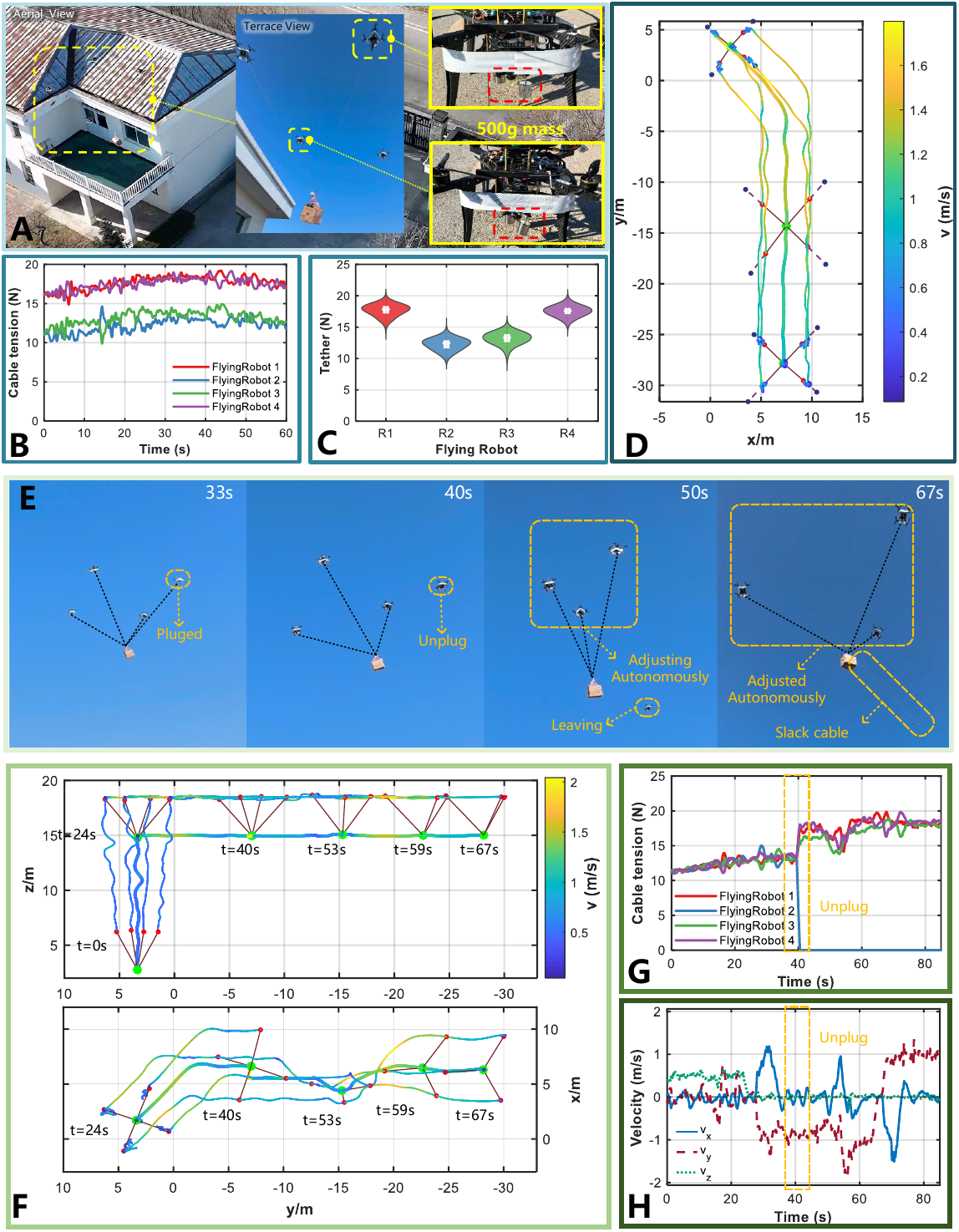}\\
		\caption{\textbf{Experiment with different payload capacities and experiment unplug-and-play with four flying robots.} 
			(\textbf{A}) The load capacities of flying robots are different due to two of them carrying counterweights.  (\textbf{B}) The change of cable tension of each flying robot with different load capacity over time. (\textbf{C}) The distribution of the cable tensions during the whole transportation task under the circumstance of different load capacities. (\textbf{D}) The top view of positions and trajectories of virtual nodes, flying robots and the load in the experiment with different payload capacities. (See Movie S3 in Supplementary Materials)
			(\textbf{E}) States of the flying robots, cables, and load at representative moments during one flying robot unplugged and the others adjusted autonomously.
			(\textbf{F}) The side view and top view of positions and trajectories of virtual nodes, flying robots and the load in the experiment of unplug-and-play.
			(\textbf{G}) The change of cable tension of each flying robot during the unplug-and-play experiment over time.
			(\textbf{H}) The change of the velocity of one flying robot during the unplug-and-play experiment over time.
			(See Movie S4 in Supplementary Materials)}
		\label{fig:6}
	\end{figure}
	
	This real-world flight experiment evaluates the system's robustness to variations or uncertainties in the payload capacities of individual flying robots. Such robustness is advantageous for two reasons: (1) it allows for the retrofitting of flying robots with varying specifications to meet the perception requirements of the proposed method, facilitating their participation in cooperative transportation tasks; (2)  it accommodates changes in the payload capacity of the same robots under different operational conditions, such as fluctuations in battery weight or voltage. By ensuring resilience to these payload variations, the system ensures reliable performance across a wide range of scenarios.

	As shown in Fig.~\ref{fig:6}A, four flying robots collaboratively transport a 3 kg load to the target terrace, with Robot 2 and Robot 3 each carrying a 500 g counterweight. From the swarm configuration plot in Fig.~\ref{fig:6}D, the system's state throughout the entire transportation process can be observed. Fig.~\ref{fig:6}B and Fig.~\ref{fig:6}C highlight the variations in the tension exerted by the robots on the load via the cables. Specifically, Robots 2 and 3, burdened with the 500 g counterweight, experience a reduction in their lifting capacity, resulting in a decreased tension exerted on the load. Driven by the method proposed in this paper, the system successfully completes the transport task with smooth execution, demonstrating that the proposed approach is robust enough to accommodate varying load capacities.
	
	The tracking error distributions in Fig.~\ref{fig5}B reveal that the dissipative structure achieves markedly robustness under different payload capacities, maintaining a median error of 1.8 m. In contrast, the formation-based and payload leader structures exhibit median errors of approximately 10 m and 14.5 m, respectively. The range of the tracking error of the dissipative approach represents 6.7$\%$ and 25$\%$ of the corresponding ranges for the formation-based and payload leader structures, respectively, underscoring its resilience to system heterogeneity.
	
	Fig.~\ref{fig5}C demonstrates the distribution of the tracking error of the proposed dissipative structure, formation-based structure, and payload leader structure on flying robots of different load weights. The dissipative structure maintains consistently low tracking errors across varying payloads, with median values remaining below 1 m and minimal variance. In contrast, the formation-based structure exhibits moderately increasing tracking errors and larger variance as payloads increase, while the payload leader structure shows substantial performance degradation, with median errors exceeding 2 m and mean errors approaching 20 m at 8 kg. These results demonstrate that the dissipative structure offers significantly enhanced resilience and consistency compared to the alternative methods under increasing payload.
	\paragraph*{Unplug-and-Play Flying Robots and with limited perception capability}
	
	
	
	In real-world scenarios, the failure of a single flying robot during transportation could significantly impact the mission. This risk increases exponentially as the number of flying robots involved increases. The proposed cooperative transportation method handles such failures effectively by redistributing the workload among the remaining robots. This real-world flight experiment demonstrates the system’s capability to dynamically remove flying robots during transportation. Once a failed robot exits the perception range of the others, the system autonomously readjusts, ensuring stable operation.

	In this real-world flight experiment, four flying robots are employed to transport a 4 kg load, with each robot equipped with a release mechanism to disengage from the cable. From Fig.~\ref{fig:6}E and Fig.~\ref{fig:6}F, it can be observed that before 40 s, all four robots are connected to the load. At the 40th second, Robot 2, enclosed by the yellow dashed line, actively disengages and disconnects from the load, then distances itself from the system.  After 40 s, the remaining three robots, guided by the proposed method, autonomously adjust their distribution and reorganize their formation, ultimately adopting a new configuration resembling an equilateral triangle. This arrangement was most conducive to the even distribution of cables of the load. As observed in Fig.~\ref{fig:6}G, after Robot 2 disengaged at 40 s, its tension dropped rapidly to zero, while the cable force of the remaining three robots surged and adjusted swiftly. After 13 s, the adjustment was completed, and the tension distribution became relatively balanced. Fig.~\ref{fig:6}H shows the velocity of Robot 1 throughout the transportation process, showing no significant fluctuations, thereby highlighting the smooth control achieved through the proposed method. In conclusion, the method proposed in this paper effectively manages collaborative transport tasks in the event of robot failure and voluntary disengagement, thereby enabling hot-swapping functionality.
	
	Based on the boxplot presented in Fig.~\ref{fig5}D, the tracking error of the dissipative structure demonstrates a markedly lower variability compared to the formation-based structure. Specifically, the dissipative structure consistently exhibits smaller tracking errors across different perception ranges (0.5 m, 3 m, and 8 m), with the median values remaining close to zero. In contrast, the formation-based structure shows more significant tracking errors, especially at larger distances. This highlights the robustness of the dissipative structure to variations in perception range, making it more resilient in maintaining low tracking errors. 
	
	\section*{Discussion}
	
	The proposed cooperative transport method exhibits several key advantages over existing solutions: generalization to heterogeneous flying robots, adaptability in communication-denied environments, insensitivity to imprecise information, and robustness to partial failures during operation. By employing LiDAR-based relative localization, flying robots are able to dynamically adjust the entire formation through an interactive force among neighbors, obviating the necessity for explicit communication. This enables the robot swarm to function in a communication-denied environment. Moreover, the load is automatically distributed to heterogeneous flying robots with different payloads by regulating corresponding virtual nodes. The flying robot swarm is also readily scalable without requiring a redesign of the overall architecture, as the method does not need the pre-design of a formation based on the explicit number of flying robots.
	Additionally, the mechanism exhibits a high tolerance for partial participation or failure of flying robots, as the dissipative feature has guaranteed the overall system stability.
	
	In order to more accurately assess the adaptability, scalability, and robustness of the dissipative transportation system, this paper compares it with two well-established transport strategies under various scenarios. One approach is based on centralized control, which involves configuring a fixed formation based on the position of the suspended load. Specifically, the position of the load and the formation configuration are employed to calculate the expected position of each flying robot during the transport procedure, with trajectory control conducted on an individual basis. Another approach is to maintain a pre-designed formation with an explicit connection among flying robots. This formation control allows flying robots to maintain a stable distance between each other, thereby ensuring the stability of the transported load. In the typical case of point-to-point cargo transportation, the above two approaches are compared with the proposed method under conditions of uncertain information, different payloads, and limited vision. The overall performance of the three methods is also evaluated when the aforementioned disturbances and uncertainties exist simultaneously. A detailed benchmark is presented in Supplementary Materials.

	\paragraph*{Generalization to heterogeneous flying robots}
	
	Most of the current cooperative transportation methods require pre-designing swarm formation based on the type and number of flying robots. However, this results in a less flexible structure when a robot joins or departs. The dissipative transportation structure adaptively adjusts the number of flying robots and the formation through virtual nodes, thereby facilitating a flexible solution for cooperative transportation.  As is common with many physical systems observed in nature, the dissipative system favors a uniform distribution of the lowest energy states, rather than concentrating them in a specific location. Upon adding new robots to the transportation system, the system will achieve a new equilibrium through interaction and reform another uniform distribution under the same framework. Specific simulation results with varying numbers of flying robots can be found in S5 in Supplementary Materials. It is noteworthy that a larger safety distance among robots is required for safety after scaling up. Variation of the coefficients of virtual nodes is also required within heterogeneous robot swarms. A movie of 500 flying robots' cooperative transportation is presented in Movie S7 in Supplementary Materials.

	\paragraph*{Stability with different uncertainties}
	
	In real-world scenarios, the information regarding cable length, load position, and inter-robot distances is inherently imprecise due to sensor measurement errors. Consequently, these uncertainties introduce uncertainty in the transportation process. Since the suspending force is adaptively adjusted based on the distance changes between virtual nodes and their corresponding positions of the flying robots, the dissipative transportation structure remains resilient to uncertainties in cable length and is validated in S6 in Supplementary Materials, demonstrating the smallest tracking error. The measurement uncertainty of the internal distances within the formation represents another source of imprecise information. While the deformation of virtual springs is influenced by these measurements, it does not significantly compromise the transport stability. This is due to the dissipative characteristics that ensure the stability of the transportation structure, which is validated by real-world flight experiments in the Results section.

	\paragraph*{Adaptability in communication-limited environment}
	
	Explicit communication is a critical requirement in many transportation methods, as flying robots depend on the positions of neighboring robots to maintain precise distances. However, establishing a reliable communication link is challenging due to potential obstructions and interference during transportation. The payload leader structure functions in an environment lacking communication and is more resilient when confronted with communication disruptions. However, this method is less flexible and can only accommodate a fixed number of flying robots. The dissipative structure emulates collaborative transport involving multiple human workers, thereby ensuring resilience in the event of a communication-denied environment. In instances where a worker is transporting a substantial load, the worker's view may be obstructed, potentially impeding the ability to navigate the environment safely and effectively. Nevertheless, this does not significantly compromise stability, as the worker is able to adjust force and position based on the remaining visible workers. During the transmission process, flying robots' perception ability is constrained by the onboard LiDAR's limitations. The dissipative forces are maintained within the robot's perception range, analogous to the visual capabilities of human beings. 
	
	\paragraph*{Robustness in halfway failure}
	
	In the context of long-distance transportation, it is critical to ensure the system's resilience against the potential failure of robotic components. Transportation structures in nature typically demonstrate robust performance when members are added or removed. The dissipative transport structure employs the flexibility of virtual springs to achieve this robustness. In the event of a robot failure, the affected cable is disconnected. As other robots assume the role of transport, the failed robot will no longer be within the perception range of the remaining robots. The connection matrix is then modified according to the perception range, and the transport structure is reorganized to account for the remaining robots.
	In the failure analysis presented in S8 in Supplementary Materials, it is observed that by introducing a generalized dissipative force within the transport system, the dissipative system is able to achieve an adaptive structure, thereby enabling the robots to respond in a natural manner when faced with failures. 
	A movie of a real-world flight experiment with halfway failure is presented in Movie S5 in Supplementary Materials.
	\paragraph*{Limitations and future directions}
	From a hardware perspective, the power system of the flying robot used in the experimental section of this paper is insufficient for carrying loads of hundreds of kilograms. To extend the application of the transport system proposed in this paper to fields such as construction and emergency rescue, the power and propulsion systems of the flying robots should be upgraded while retaining other components such as sensors and control systems. For example, a fleet of flying robots with a 20 kg payload capacity could collaborate to transport loads up to 100 kg. Additionally, the current design relies on a Differential Global Positioning System (DGPS) for self-localization. In the future, optimization of sensor configurations, such as LiDAR and cameras, along with the development of related fusion and localization algorithms, could enable the collaborative transport system to operate independently of external navigation information.
	
	At the algorithmic level, the dissipative cooperative system addresses several key real-world challenges, such as scaling up robot swarms, managing the uncertain length of the cable, dealing with limited perception, and handling halfway failures. Nonetheless, more complex and dynamic environments should be considered to support a wider range of transport scenarios. By detecting terrain features and obstacles during flight, the system can adapt its trajectory in real time to avoid potential hazards such as uneven ground, sudden elevation changes, trees, buildings, or power lines. In addition to environmental awareness, active control of the suspended payload's pose is critical for minimizing oscillations and maintaining stability throughout transport. A key advantage of the proposed dissipative framework lies in its inherent scalability and resilience. These features enable the system to seamlessly adapt to missions of varying complexity, from small-scale deliveries to large, distributed transport tasks, while maintaining stable operation in the face of unexpected disruptions. Together, these capabilities position the system as a flexible and reliable solution for real-world cooperative aerial transportation.
	
	\section*{Material and methods}
	
	\subsection*{Robot Platform of the Cooperative Transportation System}
	The flying robot platform is a quadcopter equipped with an omnidirectional LiDAR, a fisheye camera, a GNSS receiver, an onboard computer, and an autopilot, as shown in Fig.~\ref{fig:UAV1}. Self-localization is achieved through the GNSS receiver. To enable relative localization without inter-robot communication, reflective tape is affixed to each robot \cite{10816004}, allowing the omnidirectional LiDAR to detect nearby agents with high positional accuracy. Payload position and state estimation are performed using the fisheye camera, which captures images of the suspended load. A combination of an object detection algorithm and a pinhole imaging model provides robust and reliable visual localization of the payload \cite{redmon2016you}. Additional configuration details are provided in Supplementary Materials S1. The suspension system uses nylon braided cords, chosen for their high tensile strength and minimal elongation. Payloads are connected via S-shaped carabiner hooks, selected for their strength, durability, and ease of attachment and release. Each quadcopter has a stable payload capacity of 2 kg.

	\subsection*{LiDAR and vision-based perception and localization}

	During cooperative transportation, the dissipative method requires flying robots to possess recovery capability—that is, the ability to perceive the real-time relative positions of neighboring robots and the payload, and to use this feedback to continuously adjust their expected positions.
	To solve this problem, a target perception and positioning system based on multiple panoramic sensors is proposed. The system utilizes a fisheye camera with a field of view of 210° and a MID360 three-dimensional LiDAR sensor. 
	The sensor installation is shown in Fig.~\ref{fig:UAV1}. Both sensors feature a 360° horizontal field of view, enabling them to capture complete environmental information in all directions within each sampling cycle. This allows the system to simultaneously acquire the position of all detectable targets in the current scene \cite{zhu2018downside}.
	During transport, the payload is typically suspended beneath the flying robots, and the altitude of each robot in the swarm remains nearly identical. Leveraging this spatial arrangement, the fisheye camera is used to detect and localize the payload, while the LiDAR sensor is tasked with acquiring the positions of neighboring flying robots within the swarm. Specifically, the LiDAR extracts surrounding point cloud data within a 2 m
	radius using distance and height-based filtering. To suppress noise and improve detection accuracy, a point cloud clustering algorithm based on a K-dimensional Tree is applied to segment the data into distinct clusters \cite{shen2021radar}. Each cluster is assumed to represent a nearby robot and is tracked over time using an Extended Kalman Filter (EKF).
	
	Since LiDAR sensors have blind spots in the vertical direction, fisheye cameras are used to detect and locate the suspended payload. Firstly, target detection is conducted to extract load information from the fisheye image. Due to the vertical blind spots inherent in LiDAR sensors, a fisheye camera is employed to detect and localize the suspended payload. The process begins with target detection to extract payload information from the fisheye image. However, the significant radial distortion in fisheye imagery makes conventional feature extraction techniques ineffective for capturing consistent load features from multiple viewing angles \cite{liu2016feature}. To address this challenge, a deep learning–based object detection method is adopted to robustly identify the payload \cite{redmon2016you}. This method delivers strong real-time performance and high detection accuracy on embedded computing platforms, effectively generating a bounding box of the target. Secondly, the center of the detected bounding box is used as a reference point, and the projection model of the fisheye camera is applied to extract the azimuth vector from the camera to the target center. To estimate depth, a fourth-order polynomial fitting algorithm is used to model the relationship between the target's distance to the camera and the diagonal length of the detection box, based on known object dimensions. The ratio between the actual payload size and the calibrated reference target is used as the correction factor to refine the distance estimation between the target and the camera. Finally, the relative distance between the payload and the flying robot is determined by combining the estimated distance from the camera to the payload with the known spatial relationship between the camera and the robot. To improve the accuracy of the position estimation of the suspended load, a Kalman Filter is employed to smooth and refine the results. When the load is in the LiDAR perception range, the camera orientation information is used to obtain a point cloud cluster describing the position of the load, and the point cloud cluster information is then used to update the image-based position estimation results, significantly enhancing localization accuracy.
	
	\subsection*{Dissipative Cooperative Transportation System}
	
	\subsubsection*{System Overview}
	Inspired by the structural properties of a table, we have developed a cooperative aerial transportation system whose dynamics are rooted in the dissipative behavior observed in table support structures. In this framework, each flying robot operates fully autonomously, equipped with onboard sensors that endow it with local perception capabilities. These sensors allow the robot to gather real-time state information about nearby agents and the payload, enabling coordination without the need for explicit communication. Virtual nodes are introduced in the model for load transportation and formation maintenance.  A dissipative control law, based on the principles of dissipation, ensures overall formation stability \cite{brogliato2007dissipative}. This control law dynamically adjusts each robot’s position based on local feedback, allowing the swarm to adapt to dynamic and uncertain environments. Trajectories are generated independently by each robot, which empowers the system to address a wide range of real-world challenges, including robot failures, limited perception ability, uncertainties in cable lengths, and varying swarm sizes. 
	\subsubsection*{Dissipative Transportation System Modeling}
	The cooperative transportation system is described as a robot system consisting of $n$ flying robots and a suspended load. Each robot is modeled as a double integrator. The connection points of the flying robots and the load are located at the center of mass. Since the elastic deformation of the cable is insignificant, the cable is treated as being in a stressed state. The dynamics of the robot system are described by the Lagrange equation  \cite{fantoni2012non} as 
	\begin{equation}
		\frac{\textnormal{d}}{{\text{dt}}}\frac{{\partial L}}{{\partial {{{\bf{\dot x}}}_i}}} - \frac{{\partial L}}{{\partial {{\bf{x}}_i}}}
		= {{\bf{u}}_i},i = 1,2,...,n+1
		\label{eq:3}
	\end{equation}
	where ${\bf{x}}_i$ are the coordinates of the $i$th flying robot; ${\bf{x}}_{n+1}$ are the coordinates of the suspended load; ${L}$ is the Lagrange function of the cooperative robot system; and ${{\bf{u}}_i}$ is the control input of the $i$th flying robot. In this paper, we design a dissipative cooperative transportation system to derive a generalized control law for each flying robot based on energy dissipation  \cite{brogliato2007dissipative}. Spatial coordinates of virtual nodes are defined as $\mathbf{q}_i\in\mathbb{R}^3$. Each virtual node is associated with a corresponding flying robot. These virtual nodes generate dissipative forces between neighboring nodes, guiding the overall system toward a stable, low-energy configuration. The corresponding dissipative node system in ~Equation~\ref{eq:3} is described as
	\begin{equation}
		\frac{\mathrm{d}}{\mathrm{dt}} \frac{\partial L}{\partial \dot{\mathbf{q}}_i}-\frac{\partial L}{\partial \mathbf{q}_i}+\frac{\partial \psi}{\partial \dot{\mathbf{q}}_i}=0, \, i = 0,1,...,n + 1
		\label{eq:dynamic function}
	\end{equation}
	where $\mathbf{q}_0$ is the formation center of flying robots; $\mathbf{q}_i$ is the coordinates of the $i$th virtual node and it is calculated as
	\begin{equation}
		\mathbf{q}_i=\mathbf{x}_i+\frac{h_c-h_i}{h_i-h_{n+1}}\left(\mathbf{x}_i-\mathbf{x}_{n+1}\right),  \, i = 1,2,...,n,
	\end{equation}
	$\mathbf{q}_{n+1}$ is the coordinates of the suspended load. $h_c$, $h_i$, and $h_{n+1}$ are the pre-designed altitude, the altitude of the $i$th flying robot, and the altitude of the suspended load, respectively. $\psi$ in ~Equation~\ref{eq:dynamic function} is a dissipative function,
	which will be defined in ~Equation~\ref{eq:f1} and ~Equation~\ref{eq:f2}. It should be highlighted that dissipation exists in this virtual node system, resulting in a gradual decay of the overall energy.  
	The general description of the dissipative force is
	\begin{equation}
		\mathbf{f}_i=\mathbf{h}_i\left(\mathbf{q}_i, \mathbf{x}_i, \mathbf{x}_j, \dot{\mathbf{q}}_i, \dot{\mathbf{x}}_i, \dot{\mathbf{x}}_j,\mathbf{q}_0, \dot{\mathbf{q}}_0\right), i=1,2, \ldots, n.
		\label{eq:fi}
	\end{equation}
	Based on ~Equation~\ref{eq:fi}, two typical dissipative forces are designed for energy dissipation in the node system in ~Equation~\ref{eq:f1} and ~Equation~\ref{eq:f2}. It should be noted that the total energy of the system dissipates when there is a relative motion among flying robots and that there is a positive correlation between the dissipation and the intensity of motion. Dissipative force is generated within the perception range of the onboard sensors of robots as
	\begin{equation}
		\mathbf{h}_i\left(\mathbf{q}_i, \mathbf{x}_i, \mathbf{x}_j, \dot{\mathbf{q}}_i, \dot{\mathbf{x}}_i, \dot{\mathbf{x}}_j, \mathbf{q}_0, \dot{\mathbf{q}}_0\right) = \left\{ \begin{array}{l}
			\mathbf{h}_i\left(\mathbf{q}_i, \mathbf{x}_i, \mathbf{x}_j, \dot{\mathbf{q}}_i, \dot{\mathbf{x}}_i, \dot{\mathbf{x}}_j, \mathbf{q}_0, \dot{\mathbf{q}}_0\right){\rm{,  }}\,{\left\| {{{\bf{x}}_i} - {{\bf{x}}_j}} \right\|_2} < r\\
			\mathbf{0}{\rm{,    \,\,\,                    otherwise}}
		\end{array} \right., \, i,j = 1,2,...,n.
	\end{equation}
	where $r$ denotes the perception range of flying robots. The detailed modeling of the dissipative system can be found in Supplementary Materials S2, and its features are discussed in Supplementary Materials S3. Based on the design of dissipative force, the dynamics of the dissipative node-robot-load system are further described as
	\begin{equation}
		\mathbf{M}\left( \ddot{\mathbf{q}},\ddot{\mathbf{x}}\right)=\mathbf{U}.
		\label{eq:model_main}
	\end{equation}
	
	\subsubsection*{Controller Design}
	A dissipative controller is designed to address the practical challenges encountered in cooperative transportation based on the dissipative node-robot-load system in~Equation~\ref{eq:model_main}. Each robot employs a control law that manages both load transportation and formation maintenance. The dissipative control law is designed as
	\begin{equation}
		\begin{aligned}
			\mathbf{u}_i= & -k_{d,i}k_{i j} \sum_{j=0}^{n} w_{i j}\left(\left(1-\frac{l_{i j(0)}}{l_{i j}}\right)\left(\mathbf{q}_i-\mathbf{q}_j\right)\right)_{\mathrm{hor}}+k_{i}\left(1-\frac{l_{i(0)}}{l_{i}}\right)\left(\mathbf{q}_i-\mathbf{x}_i\right)-k_{d,i}f_c\left(\dot{\mathbf{q}}_i-\dot{\mathbf{x}}_{0}\right)_{\mathrm{hor}} - {\bf{f}}_i\\
			& -k_{d,i}c_{i j} \sum_{j=0}^{n}\left(\frac{w_{i j}}{l_{i j}^2}\left(\mathbf{q}_i-\mathbf{q}_j\right)^{\mathrm{T}}\left(\dot{\mathbf{q}}_i-\dot{\mathbf{q}}_j\right)\left(\mathbf{q}_i-\mathbf{q}_j\right)\right)_{\mathrm{hor}}+\frac{c_{i}}{l_{i}^2}\left({\mathbf{q}}_i-{\mathbf{x}}_i\right)^{\mathrm{T}}\left(\dot{\mathbf{q}}_i-\dot{\mathbf{x}}_i\right)\left({\mathbf{q}}_i-{\mathbf{x}}_i\right)-m_{i}\mathbf{g}
		\end{aligned}
		\label{eq:control}
	\end{equation}
	where $k_{d,i}=\frac{{h}_{n+1} - {h}_i}{{h}_{n+1} - {h}_c}$; $k_{ij}, c_{ij}$ are stiffness and damping coefficients with regard to the $i$th and the $j$th virtual node; $k_{i}, c_{i}$ are stiffness and damping coefficients with regard to the $i$th flying robot and the $i$th virtual node; $f_c$ is the friction coefficient between the $i$th virtual node and the formation center; $w_{ij}$ is the connection coefficient which describes the perceptibility of the $i$th and the $j$th flying robot; the subscript ``\text{hor}'' denotes the value of z-axis component in this vector is zero; $l_{ij}$ is the distance between the $i$th and the $j$th virtual node while $l_{ij(0)}$ is the initial value of $l_{ij}$; ${\bf{f}}_i$ is the tension of the $i$th cable; $\mathbf{x}_0=\mathbf{q}_0$ is the formation center. Particularly, dissipative forces are modeled as
	\begin{equation}
		\begin{aligned}
			{{\bf{h}}_{1,i}} = & -k_{d,i}k_{i j} \sum_{j=0}^{n} w_{i j}\left(\left(1-\frac{l_{i j(0)}}{l_{i j}}\right)\left(\mathbf{q}_i-\mathbf{q}_j\right)\right)_{\mathrm{hor}}+k_{i}\left(1-\frac{l_{i(0)}}{l_{i}}\right)\left(\mathbf{q}_i-\mathbf{x}_i\right) \\
			& -k_{d,i}c_{i j} \sum_{j=0}^{n}\left(\frac{w_{i j}}{l_{i j}^2}\left(\mathbf{q}_i-\mathbf{q}_j\right)^{\mathrm{T}}\left(\dot{\mathbf{q}}_i-\dot{\mathbf{q}}_j\right)\left(\mathbf{q}_i-\mathbf{q}_j\right)\right)_{\mathrm{hor}}+\frac{c_{i}}{l_{i}^2}\left({\mathbf{q}}_i-{\mathbf{x}}_i\right)^{\mathrm{T}}\left(\dot{\mathbf{q}}_i-\dot{\mathbf{x}}_i\right)\left({\mathbf{q}}_i-{\mathbf{x}}_i\right),
		\end{aligned}
		\label{eq:f1}
	\end{equation}
	\begin{equation}
		{{\bf{h}}_{2,i}} = -k_{d,i}f_c\left(\dot{\mathbf{q}}_i-\dot{\mathbf{x}}_{0}\right)_{\mathrm{hor}}, i = 1,...,n.
		\label{eq:f2}
	\end{equation}
	The function ${{\bf{h}}_{1,i}}$ describes the energy dissipated among flying robots, while ${{\bf{h}}_{2,i}}$ describes the energy dissipation between the $i$th flying robot and the formation center to prevent the rotation motion of robot swarm. Acceleration command of the $i$th flying robot is described as
	\begin{equation}
		{{\bf{a}}_{{\rm{cmd}},i}} = {{\bf{a}}_{{\rm{expected}},i}} + \frac{1}{{{m_{{{i}}}}}}{{\bf{u}}_i} + \frac{1}{{{m_{{{i}}}}}}{{\bf{f}}_i} + {\bf{g}}.
	\end{equation}
	Building upon the dissipative control formulation, we now present Theorem 1, which guarantees the stability of flying robots.
	
	\begin{theorem}
		\textbf{\textit{(Stability of flying robots)}}  For $n$ flying robots that follows the dynamics in ~Equation~\ref{eq:3}, the state of flying robots converge to the invariance set $\mathcal{C}$
		under the dissipative control strategy in ~Equation~\ref{eq:control}, where
		\begin{equation}
			\begin{aligned}
				\mathcal{C}=\left\{\mathbf{x} \mid w_{ij}\left(\dot{\mathbf{x}}_i-\dot{\mathbf{x}}_j\right)^{\mathrm{T}}\left(\mathbf{x}_i-\mathbf{x}_j\right)_{\mathrm{hor}}=0, \dot{\mathbf{x}}_i-\dot{\mathbf{x}}_{0}=\mathbf{0},\left(\mathbf{q}_i-\dot{\mathbf{x}}_i\right)^{\mathrm{T}}\left(\mathbf{q}_i-\mathbf{x}_i\right)=0, i,j=1,2, \ldots, n\right\}.
			\end{aligned}
		\end{equation}
		%
	\end{theorem}
	
	According to Theorem 1, the states of the robotic system converge to the invariant set $\mathcal{C}$. Within this set, each flying robot maintains a constant distance from its neighbors and the formation center, and all robots share the same horizontal velocity. A detailed proof is provided in Supplementary Materials S4. This convergence ensures that the cooperative transport system reaches a stable formation where internal forces among robots are balanced. Furthermore, the spatial relationship between the suspended payload and the flying robots is characterized by Theorem 2, as described below.
	
	\begin{theorem}
		For the dissipative system described by virtual nodes in~Equation~\ref{eq:dynamic function}, the state of the suspended load on the horizontal plane $\mathbf{p}_{n+1}$ is in the convex hull of $n$th flying robots and is described as a convex combination of the position of  $n$th flying robots as
		\begin{equation}
			\begin{aligned}
				\mathbf{p}_{n+1} =\frac{\alpha_{(n+1) 1}}{\sum_{i=1}^{n} \alpha_{(n+1) i}} \mathbf{p}_1+\frac{\alpha_{(n+1) 2}}{\sum_{i=1}^{n} \alpha_{(n+1) i}} \mathbf{p}_2+\ldots+\frac{\alpha_{(n+1) 2}}{\sum_{i=1}^{n} \alpha_{(n+1) i}} \mathbf{p}_n.
			\end{aligned}
		\end{equation}
		Here $\mathbf{p}_i$ is the horizontal coordinates of the $i$th virtual node, and  $\alpha_{i j}=1-\frac{l_{i j (0)}}{l_{i j}}$. 
	\end{theorem}
	
	It is worth noting that $\alpha_{(n+1)i}$ is a parameter related to the size and capacity of the $i$th flying robot.If $\alpha_{(n+1)1}=\alpha_{(n+1)2}=\ldots=\alpha_{(n+1)n}$, which indicates that each flying robot has the same payload capacity, then the load is exactly at the center of the formation as
	\begin{equation}
		\mathbf{p}_{n+1}=\frac{1}{n} \sum_{i=1}^n \mathbf{p}_i .
	\end{equation}
	
	Based on Theorem 1, Theorem 2 shows that the payload's position lies within the convex hull of the virtual node positions, providing a theoretical guarantee of formation stability. The detailed proof is presented in Supplementary Materials S7. Together, Theorems 1 and 2 demonstrate that the proposed system maintains a stable formation and distributes the load effectively through virtual node coordination.
	The control law adapts to varying payload capacities and is robust against halfway failures. Despite limited sensing ranges, formation stability is achieved via decentralized dissipative control. The approach is also scalable and generalizable to diverse cooperative transport scenarios involving arbitrary numbers of robots.

		\section*{Acknowledgments}
		We thank those who offered valuable suggestions for the manuscript, the photography, and the video recording. We sincerely appreciate your help with the field experiments. \textbf{Funding}: This work was
		supported by the National Natural Science Foundation of China under Grant 62450127. \textbf{Author contributions}: Quan Quan provided the primary idea, directed the full-phase research, including the system modeling, control algorithm design, theoretical proof, simulation, and experiments, wrote a part of this manuscript, and revised the manuscript and supplementary materials. Jiwen Xu contributed to the system modeling, control algorithm design, theoretical proof, and simulation benchmark, and wrote the corresponding part of this manuscript. Runxiao Liu also contributed to improving the control algorithm, mainly by implementing the control algorithm on the hardware, testing, and iteration of the hardware platform, and real-world flight experiments, and wrote the corresponding part of this manuscript. Yi Ding contributed to the design of the perception algorithm and hardware improvement. Jiaxing Che contributed to the experimental scheme and offered the flying robot platform. Kai-Yuan Cai provided guidance and support for the research. \textbf{Competing Interests}: The authors declare that they have no competing interests. \textbf{Data and materials availability}: All data needed to support the conclusions of this manuscript are included in Supplementary Materials and at https://github.com/RflyBUAA/Cooperative-Transportation.
		

		\clearpage 

	\end{document}